\documentclass[letterpaper, 10 pt, conference]{ieeeconf}
\IEEEoverridecommandlockouts 
\usepackage{graphicx}
\usepackage{amsmath}
\usepackage{amssymb}
\usepackage{mathtools}
\usepackage{dsfont}
\usepackage{hyperref}

\usepackage[ruled,vlined]{algorithm2e}

\title{\LARGE \bf
Distributional Instance Segmentation: Modeling Uncertainty and High Confidence Predictions with Latent-MaskRCNN
}

\author{YuXuan Liu$^{1,2}$, Nikhil Mishra$^{1,2}$, 
Pieter Abbeel$^{1,2}$, Xi Chen$^1$
\thanks{$^{1}$Covariant.ai, $^{2}$University of California, Berkeley. Correspondence to: yuxuanliu@berkeley.edu. Published at ICRA 2023.}
}

\begin{document}

\maketitle
\thispagestyle{empty}
\pagestyle{empty}

\begin{abstract}

Object recognition and instance segmentation are fundamental skills in any robotic or autonomous system. 
Existing state-of-the-art methods are often unable to capture meaningful uncertainty in challenging or ambiguous scenes, and as such can cause critical errors in high-performance applications.
In this paper, we explore a class of distributional instance segmentation models using latent codes that can model uncertainty over plausible hypotheses of object masks.  
For robotic picking applications, we propose a confidence mask method to achieve the high precision necessary in industrial use cases.
We show that our method can significantly reduce critical errors in robotic systems, including our newly released dataset of ambiguous scenes in a robotic application.
On a real-world apparel-picking robot, our method significantly reduces double pick errors while maintaining high performance.

\end{abstract}

\section{INTRODUCTION}
Instance segmentation is a fundamental problem in many real-world robotic systems. 
The goal of instance segmentation is to enumerate the objects (or \textit{instances}) that appear in an image, specifying which pixels in the image belong to each object.

In the past few years, recent work has mostly focused on developing specialized architectures that make the instance segmentation task more amenable to deep learning.
For example, \textit{detect-then-segment} methods \cite{ fastrcnn, fasterrcnn, maskrcnn, rcnn, detr, zhang2022dino}, rely on a cascade of classification, regression, and filtering to first identify a \textit{bounding box} for each instance (a related problem known as \textit{object detection}), followed by an additional step to predict each instance's mask given its bounding box.
Another example is \textit{pixel-embedding} methods \cite{pixellabeling, deepwater}, which optimize pixel-level auxiliary tasks, and then use a specialized clustering procedure to extract instance predictions from the dense pixel representation.

We observe that existing methods are not well equipped to deal with the inherent ambiguity that exists in the real world.
We posit that this stems from a phenomenon we describe as limited \textit{distributional expressiveness}, namely, that most instance segmentation models are designed to predict only \textit{one} possible segmentation hypothesis (a single set of objects).
Making only a single prediction is limiting in terms of the accuracy attainable by high-performance autonomous systems: a robot picking application may only tolerate $<1\%$ of errors caused by incorrect segmentation.

 \begin{figure}[t]
   \centering
   \includegraphics[width=0.8\linewidth]{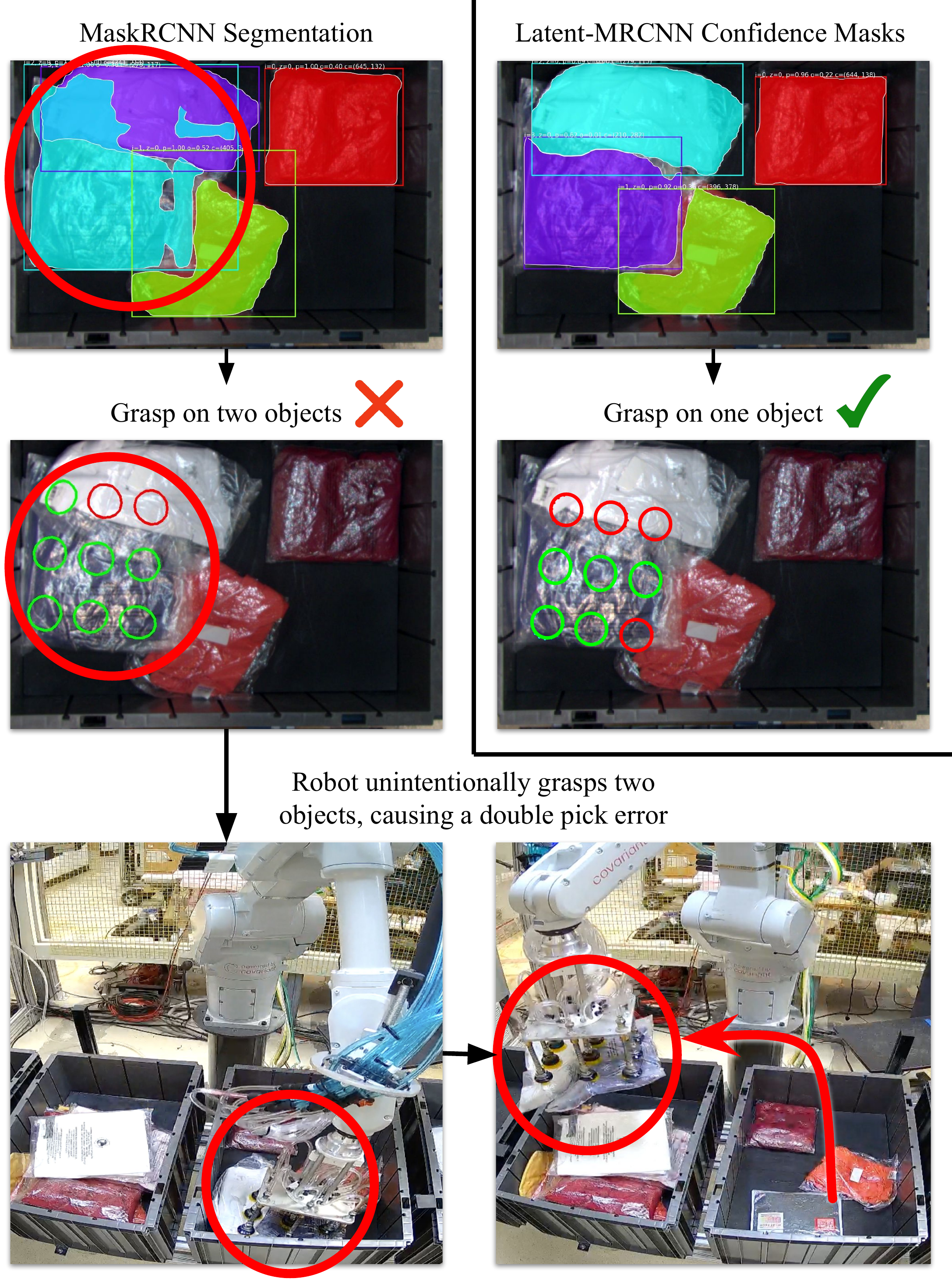}
   \caption{Traditional instance segmentation models such as MaskRCNN cannot model uncertainty over object masks. For robotics, this can result in critical errors such as unintentionally picking two objects. Our Latent-MRCNN can predict multiple hypotheses of object masks and use these to make high-confidence predictions, reducing the rate of double pick errors.}
   \label{fig:robot-overview}
 \end{figure}

To overcome these limitations, we propose \textit{distributional instance segmentation} which models a distribution over plausible hypotheses of objects. The key contributions of this work are:
\begin{itemize}
\itemsep0em 
    \item [1.]
    We introduce a distributional instance segmentation model using latent codes, Latent-MaskRCNN, which can predict multiple hypotheses of object masks.

    \item [2.]
    We propose new methods for using the output of a distributional instance segmentation model.
    For robotic applications, we propose high-precision predictions with Confidence Masks, and we achieve high recall with Union-NMS.
    
    \item [3.]
    We are releasing a dataset of over 5000 annotated images from a real-world robotics application that highlights the ambiguity in instance segmentation. We show our method achieves high performance on this dataset as well as popular driving and instance segmentation datasets.
    
    \item [4.] On a real-world apparel picking robot, our method can significantly reduce critical errors while achieving a high level of performance (Fig.~\ref{fig:robot-overview}).

\end{itemize}

\begin{figure*}
    \centering
    \includegraphics[width=\linewidth]{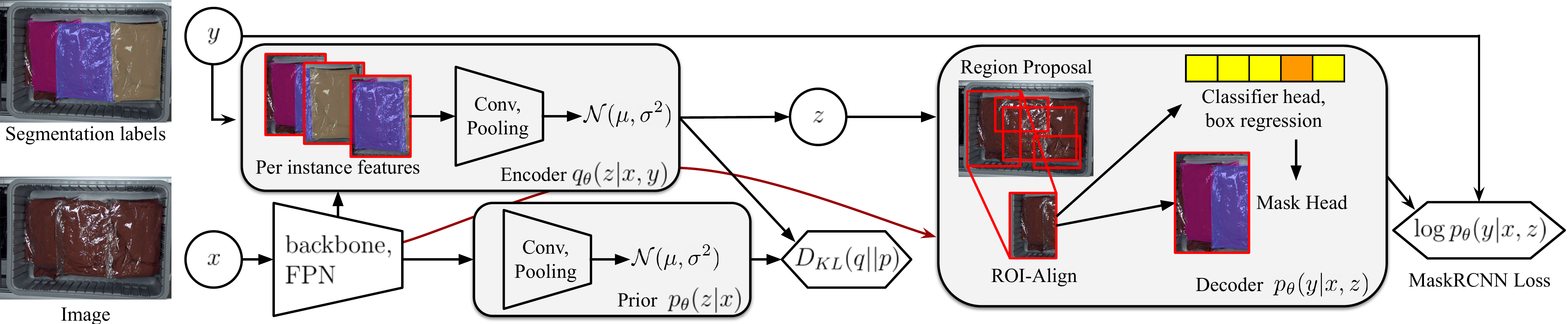}
    \caption{Overview of Latent-MaskRCNN: At training time, the encoder $q_{\theta}$ uses features extracted from the image $x$ and labels $y$ to sample a latent code $z$ which is passed into the decoder. The decoder conditions on $z$ and uses a typical MaskRCNN architecture to predict masks including region proposal, classifier, box, and mask heads. At inference time, $z$ is sampled through the prior $p_{\theta}(z|x)$ which only takes the image $x$ as input. $D_{KL}(q||p_{\theta}(z|x))$ ensures the prior has good coverage over the latent space.}
    \label{fig:arch-overview}
\end{figure*}

\section{Related Work}
\label{sec:relatedwork}

\textbf{Detect-then-segment} methods are the most popular instance segmentation methods, and MaskRCNN belongs to this category.
While they all first perform object detection and then segment each instance given its bounding box, there are some variations.
For example, YOLACT \cite{yolact} follows the same structure as MaskRCNN, but uses YOLO \cite{yolo} as the object detector instead of FasterRCNN~\cite{fasterrcnn}.
YOLO is very similar to FasterRCNN, making architectural changes that sacrifice some accuracy in exchange for real-time inference speed.
Thus, we expect YOLACT to have the same distributional limitations as MaskRCNN. 
Other methods \cite{probobjectdetection, bayesOD, sampling_object_det} explore how to express uncertainty during the detection step, but they consider distributions of individual boxes rather than over sets of object masks.

\textbf{Mask-proposal} methods \cite{tensormask, polarmask, cheng2021mask2former} aim to circumvent bounding boxes as an intermediate representation.
They are structured like FasterRCNN \cite{fasterrcnn}, but propose masks directly.
Empirically, they do not behave much differently than MaskRCNN.
Distributionally, they suffer from many of the same limitations as MaskRCNN: each proposal still models each pixel independently of the others, and they still rely on NMS to filter proposals.

\textbf{Pixel-embedding} \cite{pixellabeling, deepwater, spatial_embedding, probinstanceseg} methods work in a substantially different way than either of the above two families.
They generally optimize some auxiliary task that encourages pixels in the same instance to have similar representations.
Then they rely on a clustering-based inference procedure to extract instance predictions from their pixelwise representations.
However, their performance has lagged quite far behind that of detect-then-segment methods, which has made them relatively unpopular.
They can model per-pixel uncertainty in a manner similar to a naive semantic segmentation method, but this is likely insufficient for distributional expressiveness.

A number of methods explore how to express uncertainty in other structured prediction tasks. However, many of these do so by training multiple replicas of the entire model or some subset of the parameters, and modifying the training objective in a way that encourages diversity amongst the replicas \cite{lakshminarayanan2016simple, guzman2012multiple, rupprecht2017learning, gao2017deep}.
This incurs a multiplicative increase in the computational cost and memory footprint required at training time, which can be prohibitively expensive for large models. 
Other latent-variable formulations \cite{vae-unet, vqsegm, video_instance_segm} offer improvements on medical \textit{semantic} segmentation and video segmentation tasks.
We find, however, that \textit{instance} segmentation poses a richer set of challenges and has different application-specific uses.

\section{Distributional Instance Segmentation with Latent Variables}\label{sec:latentvar}

\subsection{Latent Variable Formulation}
\label{sec:latentvarform}

How can we turn instance segmentation into distributionally expressive models, while retaining the inductive biases of existing model architectures? Drawing on prior work in variational inference \cite{vae-unet, vae} we consider a latent-variable formulation 
where we incorporate latent codes in the style of a variational autoencoder. If we adopt this framework, then an instance segmentation model becomes a conditional VAE that is trained to maximize the evidence lower-bound:
\begin{equation}\label{equation:vae}
\log p(y|x) 
\geq \mathbb{E}_{z \sim q}[ \log p(y|x,z) ] - D_{KL}( q || p(z|x) )
\end{equation}

Typically $q(z|y,x)$ is known as the encoder, $p(y|x,z)$ as the decoder, and $p(z|x)$ as the prior, and these components are all learned to maximize the lower-bound. 
The decoder is essentially an instance segmentation model in the traditional sense, except that it is augmented to additionally consume a latent code $z$. 
This general technique allows us to reuse any existing instance segmentation model to implement our decoder (and train it in the same way), with only a slight modification to incorporate $z$ as an input.

During inference, we can sample from $p(y|x)$ by sampling different latent codes $z^{(k)} \sim p(z|x)$, and decoding them into different instance predictions $y^{(k)} \sim p(y|x,z)$.
This can be quite powerful since we can now sample multiple structured and expressive hypotheses for a given image.

\subsection{Latent-MaskRCNN}
\label{sec:latentmrcnn}

In principle, the latent-variable method can be applied to any existing instance segmentation model.
In this section, we explore how it might be applied to MaskRCNN.
We call the resulting model \textit{Latent-MaskRCNN} (Fig.~\ref{fig:arch-overview}).
We chose MaskRCNN since it is one of the most popular instance segmentation models and has served as the basis for most state-of-the-art methods in recent years.

The decoder of Latent-MaskRCNN uses the same architecture and training objective as MaskRCNN, with the main change being that it needs to incorporate latent codes.
To allow them to influence as much of the prediction as possible, we want to do this relatively early in the model.
We chose to inject the latent codes directly before region proposal, so that they can influence region proposal network, object detection head, and mask head.
We tile the latent codes across the spatial dimensions of the image and concatenate them with the feature maps from the Feature Pyramid Network (FPN) \cite{fpn}.
Then we use a few convolutional layers to project the combined feature maps back down to their original channel dimensionality.

The encoder of Latent-MaskRCNN takes in an image $x$ along with a set of ground-truth instances $y$, and produces a distribution over latent codes $q_\theta(z|y,x) = \mathcal{N}( \mu_\theta(y, x), \sigma^2_\theta(y, x) )$. 
The architecture for $\mu_\theta(y, x)$ and $\sigma_\theta(y, x)$ takes inspiration from the mask head of MaskRCNN: it acts like a "reverse mask head" that operates on each ground-truth instance, and then pools features from across all instances.
For each ground truth instance $y_i$, we extract ROI-aligned features from the FPN feature maps.
Then we use a small CNN to embed each one into a single feature vector.
At this point, we employ a graph neural network \cite{gnn} to accumulate information from per-instance features since we need to a single latent code for the entire image.
After several graph network layers, we mean-pool across the node features and use a fully-connected layer to produce a mean and log-variance for our latent distribution.
The encoder is only used at training time since it has access to the ground truth mask labels.

At inference time, we must sample latent codes from the prior to produce mask samples. This prior takes in an image $x$ and produces a distribution over latent codes $p_\theta(z|x) = \mathcal{N}( \mu_\theta(x), \sigma^2_\theta(x) )$. 
We apply a few convolutional layers to the FPN feature maps, mean-pool across the spatial dimensions, and then predict a mean and log-variance using a small MLP. For all latent distributions, we use a fixed 64-dimension Gaussian with diagonal covariance.

For training, we use the encoder $q_\theta$ to sample latent codes $z$, which are passed to the Mask-RCNN decoder. 
We maximization the evidence lower bound (Equation~\ref{equation:vae}) objective, where $-\log p(y|x,z) = \mathcal{L}_M(x, y, z)$ is the usual Mask-RCNN loss:
\begin{equation}
\mathcal{L}_M(x, y, z) = \mathcal{L}_{RPN} + \mathcal{L}_{cls} + \mathcal{L}_{box} + \mathcal{L}_{mask}
\end{equation}
$D_{KL}( q || p(z|x) )$ ensures the prior has good coverage over the encoder distribution. 
During inference, we sample latent codes from the prior (instead of from the encoder), but the decoder consumes them in the same way as during training. 

We found it helpful to use a KL warm-up, as is a common practice for training VAEs \cite{betavae}. 
The total training loss for Latent-MaskRCNN then becomes:
\begin{equation}
\mathcal{L}(x, y) = \mathbb{E}_{z \sim q}[ \mathcal{L}_M(x, y, z)] + \beta D_{KL}( q || p(z|x) )
\end{equation}
In the first part of training, we use $\beta = 0$ and increase $\beta$ towards the end of training. This allows the latent code to encode useful information early on as the rest of the model is still learning; towards the end of the training, higher $\beta$ pushes the latent space to be covered by the prior for better samples. For more details on our models and code, please refer to our website \href{https://segm.yuxuanliu.com}{segm.yuxuanliu.com}.

\section{Applying Distributional Instance Segmentation}
\label{sec:app_use_cases}

\begin{figure}[t]
    \centering
    \includegraphics[width=\linewidth]{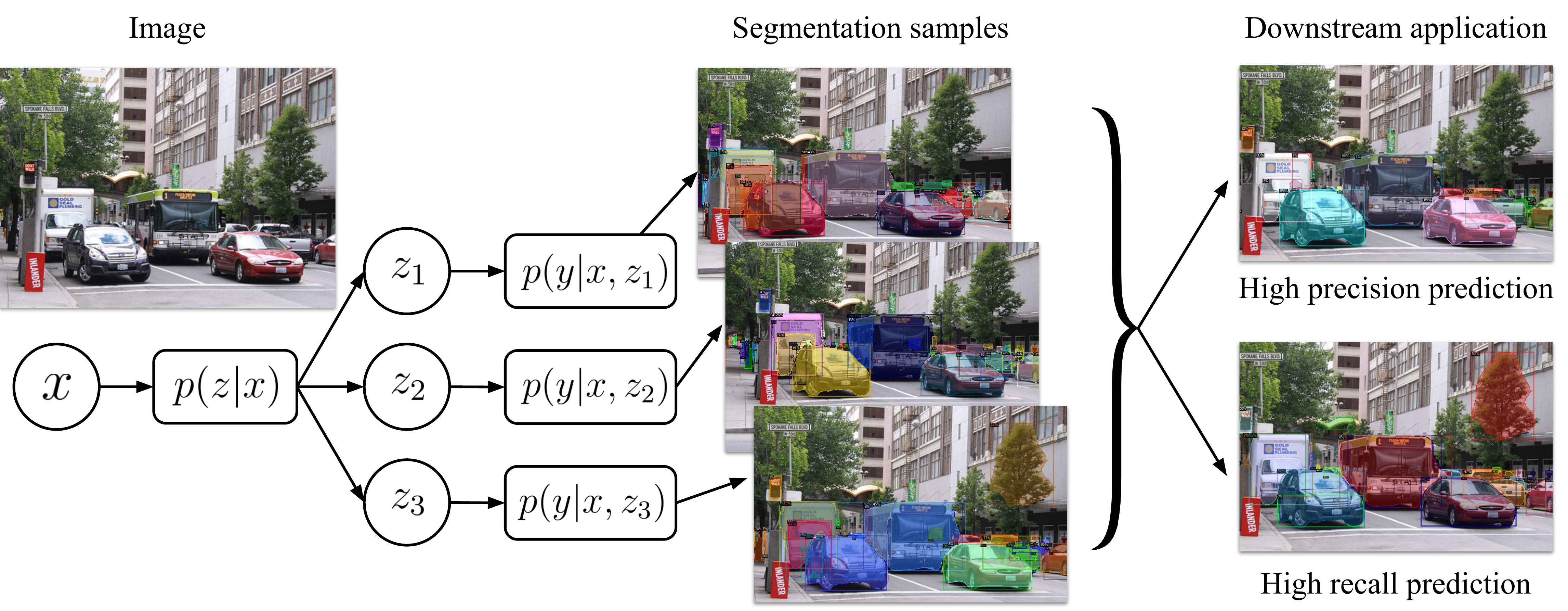}
    \caption{At inference time, the encoder $q_\theta$ is discarded, and latent variables $z_i$ are sampled from the image $x$ conditioned prior $p_\theta(z|x)$. Each latent is decoded using $p_\theta(y|x, z_i)$ into a set of masks, which can be used for our high precision or recall predictions depending on the application.}
    \label{fig:overview}
\end{figure}
 
Given a distributionally-expressive segmentation model, a natural question might be, how a downstream application can consume its distributional output?
Instance segmentation often occurs at the beginning of the perception pipeline, and it's not immediately clear how samples from a distributional segmentation model can be used downstream.
Moreover, each application may have varying error asymmetries: failing to detect an object can be catastrophic in autonomous driving but acceptable in robotic picking, while grouping two objects as one is a critical error in robotic picking but more reasonable in driving applications.
In this section, we show how a single Latent-MaskRCNN model can be used flexibly across a number of applications with different requirements (Fig.~\ref{fig:overview}).

\subsection{High-Precision with p-Confidence Masks} 
\label{sec:agreement}

In some applications, it can be very costly to make \textit{under-segmentation} errors, when an instance's mask is predicted to be larger than it actually is.
For example, consider a robotic manipulation application, where the robot must pick objects one at a time and feed them into a sortation process.
If the model undersegments an instance, it may inadvertently pick multiple objects, which can be an expensive error for the downstream application. How can we ensure that these errors don't occur?

Suppose we draw several samples from Latent-MaskRCNN.
If two pixels belong to the same instance mask in many samples, then we can be reasonably confident that they actually do belong to the same ground-truth instance.
Drawing on this intuition, we can then compute a $p$-\textit{confidence mask}, consisting of pixels that are all likely to be contained in a single ground-truth instance.

For a given confidence requirement $p$, we define a confidence mask $c_p$ as a mask that is fully contained in a ground truth mask $m$ with probability at least $p$: $\mathbb{P}(c_p\subseteq m) \ge p$. 
Using Latent-MaskRCNN, we can approximate this probability as:
$$
\mathbb{P}(c_p\subseteq m) = \mathbb{E}_{p(m|x)}[\mathds{1}\{c_p\subseteq m\}]\approx \frac{1}{k}\sum_{i=1}^{k} \mathds{1}\{c_p\subseteq m_i\}
$$
In the finite sample regime,
$\hat{c}_p$ is an empirical confidence mask if it is contained within a sampled mask for $p$ fraction of the samples.

Now consider any subset of masks $I$ consisting of one mask $m_j$ from at least $kp$ different samples. 
If we take the intersection of all of the masks in $I$, $\hat{c}_p = \bigcap_{m_j\in I}m_j$, then this intersection mask must be contained in each of the masks used in the intersection $\hat{c}_p \subseteq m_j$. Therefore we have:
$$ \frac{1}{k}\sum_{m_j \in I} \mathds{1}\{\hat{c}_p\subseteq m_j\} = \frac{|I|}{k} \ge p$$

and $\hat{c}_p$ is an empirical confidence mask by construction. 

\begin{figure}[t]
   \centering
   \includegraphics[width=\linewidth]{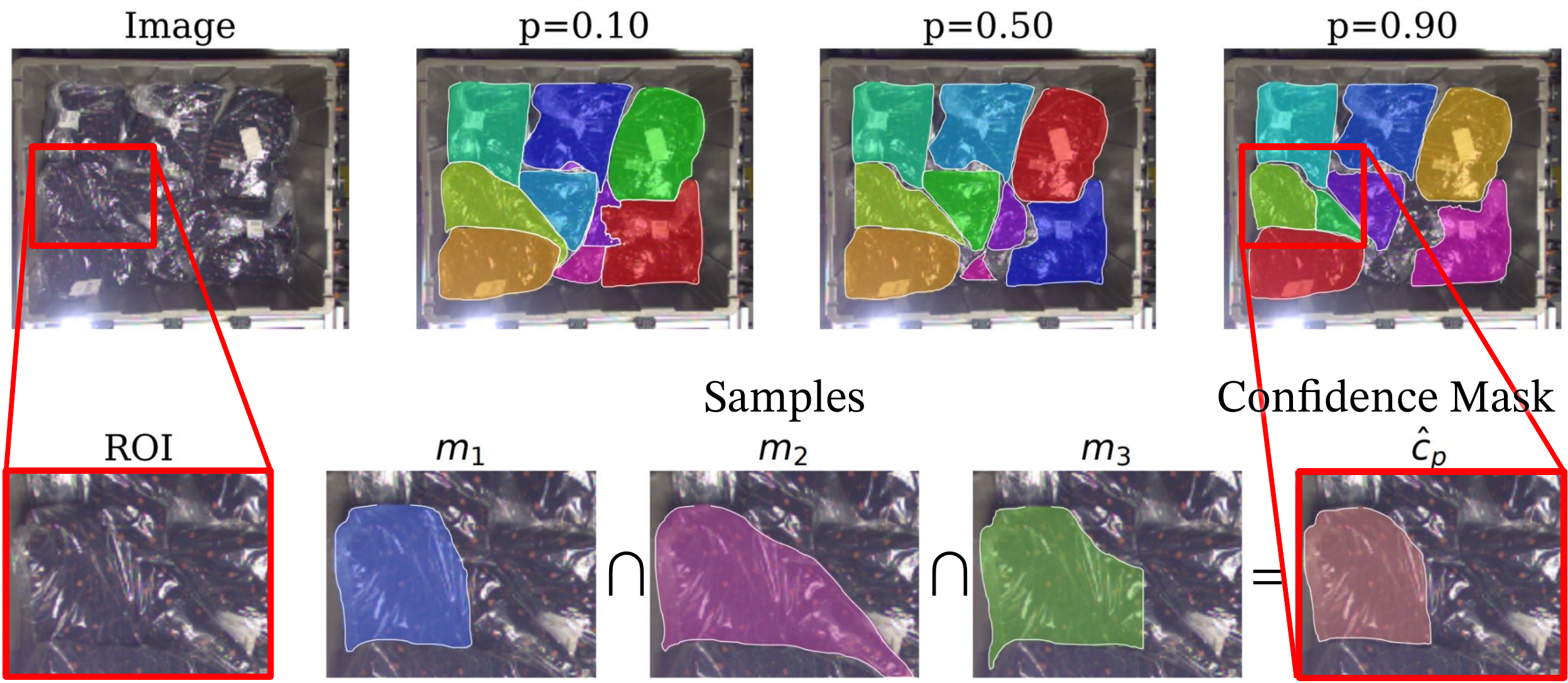}\caption{Top: Confidence Mask predictions with different $p$. Notice that as the confidence requirement $p$ increases, single objects can be split into two, ambiguous object extents are reduced, and uncertain objects are eliminated entirely. Bottom: Constructing an empirical confidence mask $\hat{c}_p$ by taking intersection of samples $m_1, m_2, m_3$. When an object's extent is uncertain, a high confidence mask prediction will only consist of pixels that are highly likely to be contained within an object as determined by the samples.}
   \label{fig:conf-mask}
 \end{figure}

Figure~\ref{fig:conf-mask} illustrates confidence mask predictions for different $p$. Notice as the confidence requirement increases, the unconfident extents of the masks shrink, and some uncertain masks are eliminated. We can also see how constructing confidence masks via intersection leads to high confidence region predictions.

\subsection{Scoring Confidence Masks}
\label{sec:score_conf}

Since each confidence mask is an intersection of masks $c_p = \bigcap_{I} m_j$, how should we assign the score of a confidence mask prediction? One intuitive approach might be to take the average of the score $s_i$ of each mask in the intersection: $\frac{1}{|I|}\sum s_j$. However, a confidence mask is not an average of masks but rather an intersection. 

To formulate a better score for our confidence mask prediction, consider two scenarios. In the first scenario, the model is very confident about an object's mask so it predicts roughly the same mask in every sample. The resulting confidence mask $c_p$ has high IoU with each of the masks used in the sample $m_j$. On the other hand, consider an unconfident prediction where the object's mask varies significantly across samples. Here, the confidence mask $c_p$ represents a small but confident region of the object whose extent is highly uncertain. The resulting IoU between $c_p$ and each $m_j$ will be smaller than when the model is confident and masks are not varying across samples.

We score each confidence mask as the mean score-weighted IoU between the predicted mask $c_p$ and every mask used in the intersection:
$$s_{c_p} = \frac{1}{|I|}\sum_{m_j \in I} s_j \frac{|c_p \cap m_j|}{|c_p\cup m_j|}$$ 
When $s_{c_p}$ is large, this indicates that $c_p$ is a confident intersection of masks with very similar IoU. On the other hand, a small $s_{c_p}$ indicates that $c_p$ has low score or IoU with its samples and likely does not capture the full extent of the object well. 

To predict a set of confidence masks for an image, we iteratively select the highest scoring confidence mask, excluding the pixels of all of the confidence masks predicted so far. 
This algorithm greedily approximates the maximum scoring confidence mask selection optimization.

\subsection{High-Recall with Union-NMS}
\label{sec:recall}

Other applications might be concerned about \textit{over-segmentation}, the complement of under-segmentation.
For example, in autonomous driving, failing to identify a pedestrian, or predicting them to be smaller than they actually are, can lead to a catastrophic error.

To make high-recall predictions with Latent-MaskRCNN, we use a procedure called \textit{Union-NMS}.
We first sample multiple segmentations from the model, and run NMS on the predicted masks. 
It checks if any two masks $m_i, m_j$ have IoU greater than some threshold, and then discards the lower-scoring one.
Suppose that some mask $m_i$ remains after we perform NMS.
Then Union-NMS returns the union of $m_i$ with every mask that it suppressed, achieving higher recall by incorporating masks that would have otherwise been ignored.

\bgroup
\def\arraystretch{1.1}
\begin{table*}[t]
  \centering
  \caption{Evaluation of MaskRCNN and Latent-MaskRCNN across various datasets and metrics.}\label{table:metrics}
  \begin{tabular}{|l|ccc|ccc|ccc|}
    \hline
  &\multicolumn{3}{|c|}{COCO} &\multicolumn{3}{|c|}{Cityscapes}  &\multicolumn{3}{|c|}{Apparel-5k}\\
    \hline
    Method  & MR@HP  & AR & mAP   & MR@HP & AR & mAP  &MR@HP&AR& mAP \\
    \hline
MaskRCNN & 20.0 &  66.0  &  35.0 & 25.3 & 55.1  &   \textbf{35.8} &  23.6  &  39.9  &  26.9 \\  
Latent Union NMS & 7.4 & \textbf{72.3} &  26.5 &  17.7  & \textbf{57.5}  &  33.6 &  13.6  &  \textbf{61.4}&  34.1  \\
Latent Confidence Mask & \textbf{22.0} & 48.1 & 30.5 & \textbf{28.0} & 48.6 & 34.1&  \textbf{42.4}  &  41.8   &  \textbf{35.1} \\
Latent Prior Mean  & 19.5 &  65.8 &  \textbf{35.3} &  25.7 & 53.8 &  35.0    &  26.9  &  49.4   &  34.3 \\
    \hline
  \end{tabular}
\end{table*}
\egroup

\subsection{Vanilla Prediction with the Prior Mean}
\label{sec:meanprior}

Some applications may not have any specific performance requirements or may have strict inference time requirements. In these cases, a point estimate can be sufficient. 
With Latent-MaskRCNN, we can achieve this by always decoding the mean of the prior: $z=\mu_\theta(x)$ where $\mu_\theta(x)$ is the mean of the prior $p_\theta(z|x)$ (for Gaussian $p_\theta(z|x)$, it is also the mode of the distribution).
We found that this scheme typically matches or yields a small improvement over MaskRCNN predictions, suggesting that Latent-MaskRCNN strictly increases the expressiveness of MaskRCNN and no performance is lost by using a more expressive distribution.

\section{Experiments}

We conducted experiments seeking to answer the following questions:
\begin{itemize}
\itemsep0em 
    \item [1.]
    Can Latent-MaskRCNN with Confidence Masks make high-precision predictions across a variety of datasets?
    \item [2.]
    Can Union-NMS make high-recall predictions?
     \item [3.]
    Can Latent-MaskRCNN reduce critical double pick errors in robotic picking applications?
\end{itemize}

\subsection{Datasets}\label{sec:datasets}

To help us answer these questions, we compared MaskRCNN and Latent-MaskRCNN across several datasets, each with its own set of challenges.

\textbf{COCO} \cite{coco}: This large dataset is the standard benchmark for instance segmentation. There are many object categories and a huge variety in image composition.

\textbf{Cityscapes} \cite{cityscapes}: A real-world dataset from an autonomous driving application. Although it is smaller and more specialized than COCO, it is still a popular benchmark for instance segmentation. One notable challenge is that there are many background instances that are still important to segment (e.g. pedestrians), but the limited image resolution can introduce some uncertainty.

\textbf{Apparel-5k}: We collected this dataset of roughly 5000 images from a robot picking application. We use 4198 images in the training set and 463 in the validation set. There is only one object category, but the images exhibit a lot of inherent ambiguity due to complex occlusions, lighting, transparency, etc. We are releasing this dataset on our website \href{https://segm.yuxuanliu.com}{segm.yuxuanliu.com} for the broader community to build upon our work.

For each dataset, we trained both MaskRCNN and Latent-MaskRCNN on 8 GPUs using MaskRCNN's released hyperparameters and training schedules.
We use the same publicly available train/val splits for all experiments and datasets.
We used a Resnet-50 backbone \cite{resnet}, initialized from pretrained-Imagenet \cite{imagenet} weights (for COCO) or pretrained COCO weights (for other datasets). 
Inference with MaskRCNN can take 80-100ms depending on the number of objects, and inference with Latent-MaskRCNN can take 500-1000ms depending on the number of samples and objects.

\subsection{Evaluating p-Confidence Masks}
\label{sec:precision}

In Section~\ref{sec:agreement}, we introduced high precision predictions with p-Confidence masks to address the problem of under-segmentation. 
In those cases, we care that predictions have high Intersection-over-Prediction: $\text{IoP}(m_i, g) = \frac{|m_i \cap g|}{|m_i|}$. When IoP is high, errors due to under-segmentation are less likely to occur.

When evaluating models in this regime, we need to trade off precision (in terms of IoP) with recall (to avoid degenerate solutions).
To do this, we consider the \textit{max recall at high precision} (MR@HP):
$$
\text{MR@HP} = \frac{1}{|p| \cdot |\tau|}\sum_{p_i \in p, \tau_j \in \tau}\max_{t: \text{Precision}(\tau_j) \ge p_i} \text{Recall}(t, \tau_j)
$$
For a given precision threshold $p_i$ and IoP threshold $\tau_j$, we can compute the max recall that each model achieves (or zero, if it never achieves precision $p_i$).
The MR@HP metric is the average of these recalls, over a range of precision threshold $p$ and IoP thresholds $\tau$.
For high precision use-cases, we care about performance at high values of these thresholds, therefore we use $p = \tau = [0.75, 0.8, 0.85, 0.9, 0.95]$.

In Table~\ref{table:metrics}, we evaluated Latent-MaskRCNN using both the prior-mean scheme from Section~\ref{sec:meanprior} as well as confidence masks with a confidence level $p=0.9$.
Across all three datasets, we find that latent confidence masks yield the best performance in terms of MR@HP. 
As for mAP, we find that Latent Prior Mean can match if not exceed the performance of MaskRCNN on all three datasets. 
On the challenging Apparel-5k dataset, we find that Latent Confidence Mask and Latent Prior Mean significantly outperform MaskRCNN in terms of MR@HP and mAP.
Overall, we find that Latent-MaskRCNN is a strict improvement over MaskRCNN by matching overall detection performance in terms of mAP and offering the best high-precision performance in terms of MR@HP.

\subsection{Evaluating Union NMS}
For the over-segmentation problem, we introduced the Union NMS method in Section~\ref{sec:recall}.
In such cases, we care that we have high recall (that we detect every instance that exists), and that each mask prediction has high \textit{IoG} (intersection-over-ground-truth): $\text{IoG}(m_i, g) = \frac{|m_i \cap g|}{|g|}$.
To capture both of these considerations, we consider the average-recall (AR) \cite{coco} using IoG.
This measures both recall while also penalizing over-segmentation (are there any predicted masks that are too small).

We evaluated Latent-MaskRCNN using both its prior-mean (Section~\ref{sec:meanprior}) and Union-NMS.
In Table~\ref{table:metrics}, we show that the prior-mean predictions are similar to MaskRCNN on all three datasets, while Union-NMS achieves substantially higher AR (using IoG). 
This suggests that Latent-MaskRCNN with Union NMS can more effectively cover different modes of uncertainty, for high-recall applications.

\subsection{Can confidence masks reduce double pick errors on Apparel-5K?} \label{section:industrial_metrics}

In a robotic picking application, it is costly for the robot to pick up two items accidentally, thinking it had only picked one item since it affects inventory counts and downstream orders. For the Apparel-5k dataset, we can estimate the double pick rate of a model's segmentation prediction by approximating the robot's gripper as a circle with a fixed radius in pixel space. Then, we randomly sample circles on the image and count the number of circles $D$ that land within one predicted mask but more than one ground truth mask. We divide this by the number of circles $N$ that land within one predicted mask, to arrive at the estimated double pick rate $R = \frac{D}{N}$. Empirically we find that this simulated double pick rate is correlated with double pick rates on a real robot.

\begin{figure}[t]
  \begin{center}
    \includegraphics[width=0.9\linewidth]{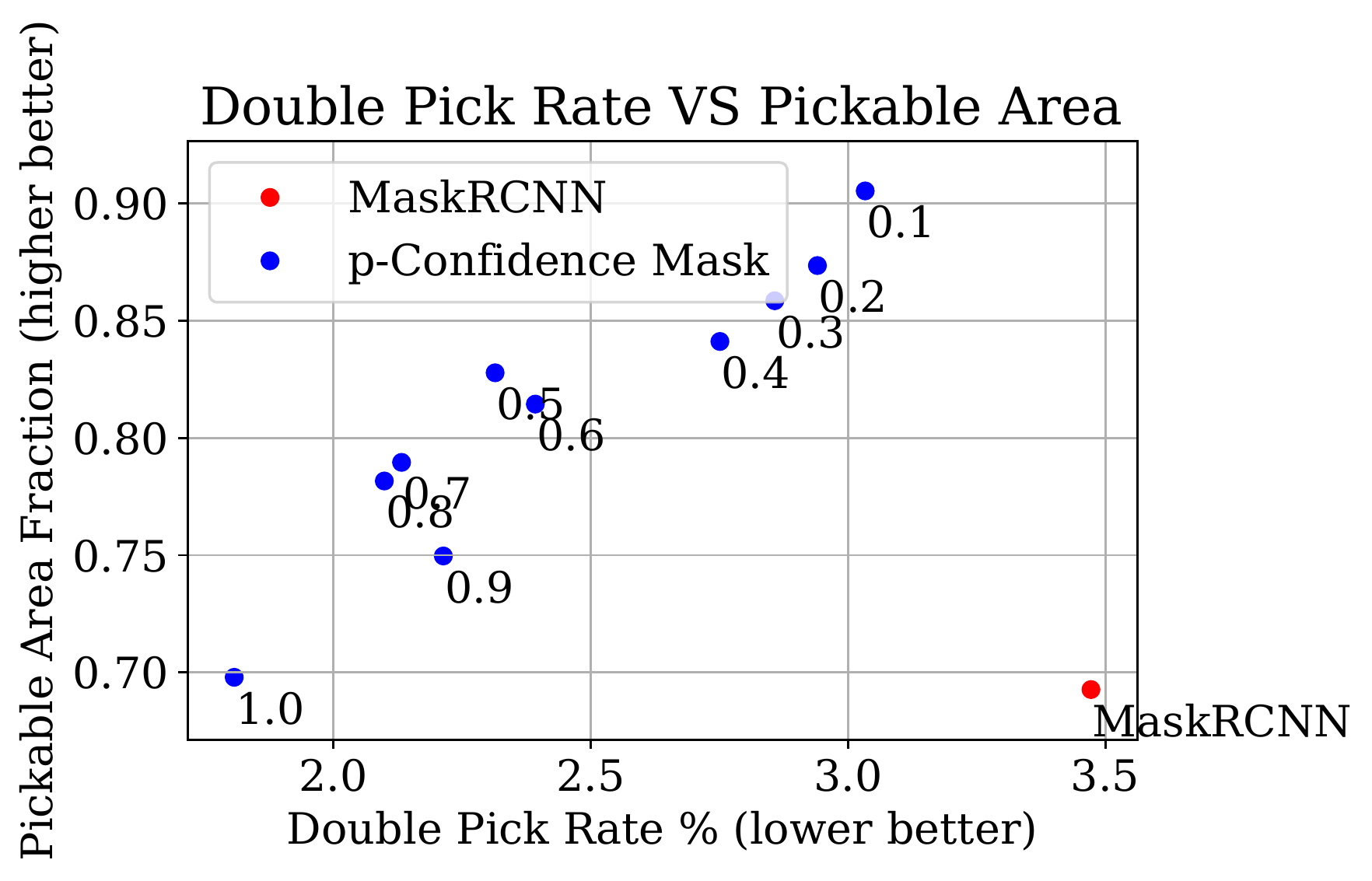}
  \end{center}
  \caption{Latent confidence masks achieve lower double pick rates and generally more pickable area compared to MaskRCNN.}
  \label{fig:doublepick}
\end{figure}

Another metric that we are concerned with in industrial robot picking is pickable area, the amount of visible surfaces that the robot can pick from. A model that predicts higher, more accurate pickable areas enables the robot to have more flexibility in its grasping strategy. To this end, we compute the area of all the predicted masks over the area of all the ground truth masks, as the fraction of pickable area available.

We compare MaskRCNN and Latent-MaskRCNN with varying $p$-confidence masks in Figure~\ref{fig:doublepick}. We find that Latent-MaskRCNN outperforms MaskRCNN in fraction of pickable area and double pick rate in all cases. Moreover, the tunable parameter $p$ in Latent confidence masks allows for application-specific tradeoffs between double pick rate and pickable area. Higher values of $p$ tend to correspond to lower double pick rate and less pickable area, as the confidence requirement for each prediction is increased. With traditional MaskRCNN, only one double pick rate and pickable area fraction is realizable since no tunable knob exists.

\subsection{Can confidence masks reduce double pick errors on a real-world apparel-picking robot?}

To evaluate whether our dataset evaluation translates to real-robot performance, we compare MaskRCNN and Latent-MaskRCNN on an apparel-picking robot. We use an ABB1300 with a 9-cup suction gripper to pick apparel items in polybags between two totes (Fig.~\ref{fig:robot-overview}). The robot uses two overhead camera systems to perform instance segmentation and then grasp point generation. The grasp points are optimized to land as many suction cups as possible on a single object detected by the segmentation model. 

For our evaluation, we only change which segmentation model is used while holding other parts of the system constant, including hardware, object set, and grasp point generation. Each segmentation model is trained on the same Apparel-5k dataset. We run each model with several hundred grasps and record the number of double picks, grasps that unintentionally pick two objects. In an industrial warehouse application, these double picks are very costly errors since they result in incorrect inventory counts and cause errors in downstream sortation and order fulfillment systems. A typical high automation warehouse can tolerate at most 1\% double pick rate before the robot is causing more problems than it solves.

We also measure the average number of sealed cups on a grasped item, since the suction holding force is proportional to the number of sealed cups. Grasps that use less sealed cups tend to result in more dropped objects, which leads to jams, lost inventory, and costly human intervention. A robotic system can reduce double pick rate by shrinking object mask sizes, chopping up bigger masks into smaller ones, or only using a single small suction cup. However, all of these approaches indiscriminately reduce the suction holding force on all items, whereas our approach will be conservative only when ambiguity is present.

Table~\ref{table:robot-metrics} reports the results of our apparel-picking experiments. We find that Latent-MaskRCNN with 0.9-Confidence mask significantly reduces the double pick rate. This validates our simulated findings on Apparel-5K in Section~\ref{section:industrial_metrics}. Moreover, Latent-MaskRCNN achieves slightly better average number of sealed cups, suggesting that suction stability was not sacrificed. This suggests that our method can make high-confidence predictions and make the appropriate trade-offs in the face of uncertainty.

\begin{table}[t]
  \centering
  \caption{Apparel-picking robot evaluation. $^*$ indicates a statistically significant difference}\label{table:robot-metrics}
  \begin{tabular}{|l|c|c|}
    \hline
    Method & Double Pick Rate & Average Sealed Cups \\
    \hline
    MaskRCNN & 4.40$\%^*$ & 4.76 \\
    Latent-MRCNN & \textbf{0.82$\%^*$} & \textbf{4.91} \\
    \hline
  \end{tabular}
\end{table}

\section{Discussion}
\label{sec:conclusion}

We proposed a new family of models that builds on top of existing instance segmentation models by using latent variables to achieve more distributional expressiveness.
Latent-MaskRCNN can express a wide range of uncertainty where existing instance segmentation models often fall short. 
We can leverage uncertainty expressed by the model using Confidence Masks and Union-NMS to achieve high precision and high recall respectively.
These methods demonstrate strong performance across robotics, autonomous driving, and general object datasets.
On a real apparel-picking robot, we find that our model can significantly reduce the rate of critical errors while maintaining high performance.
Finally, we have highlighted the importance of distributional expressiveness and hope that future work in instance segmentation can continue to build on top of our work and datasets shared in this paper.

\bibliographystyle{IEEEtran}
\bibliography{references}

\newpage
\section*{APPENDIX}

\subsection{Latent-MaskRCNN Architecture and Training}
\label{appendix:arch}

\begin{figure}
    \centering
    \includegraphics[width=\linewidth]{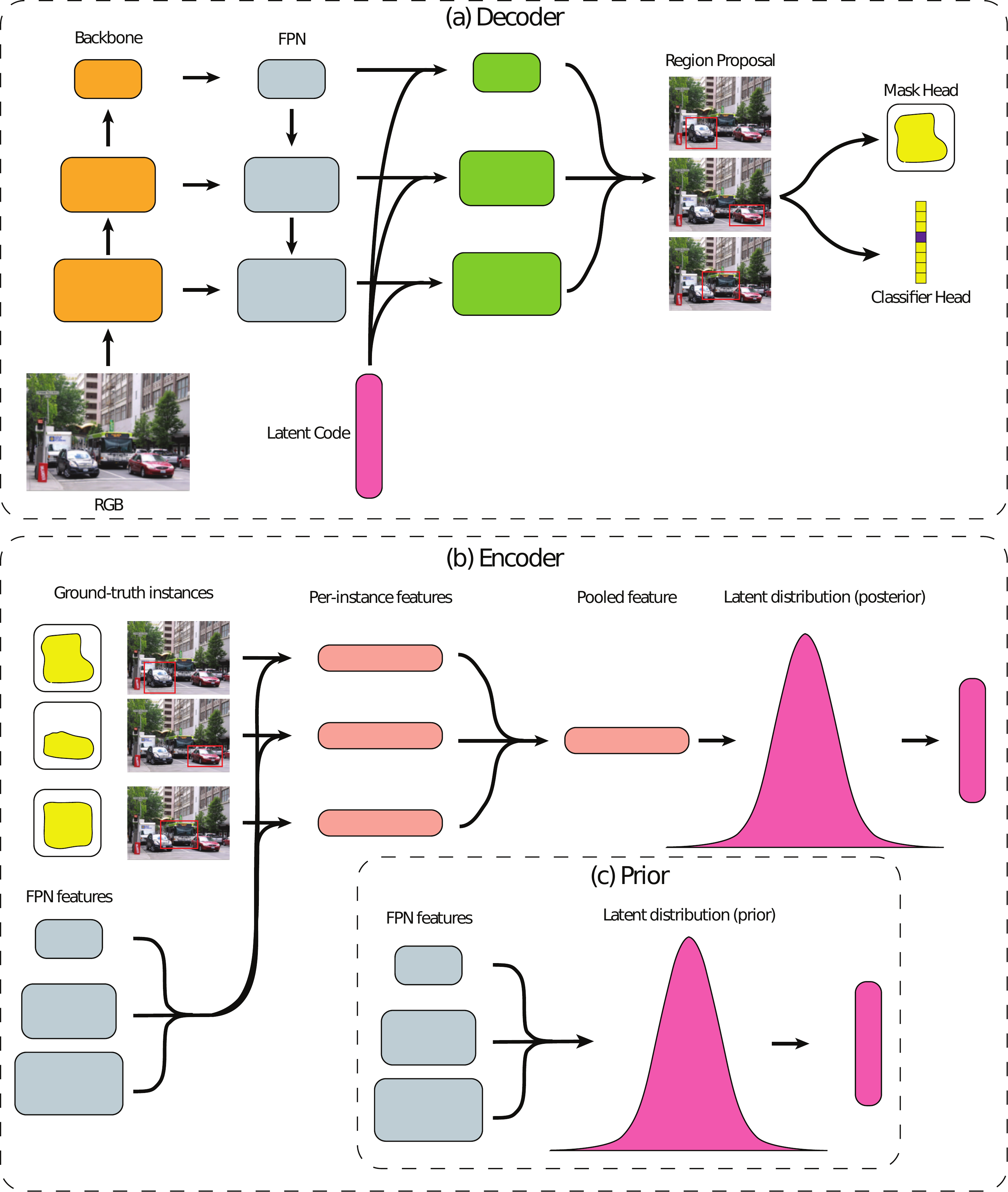}
    \caption{
    The architecture of Latent-MaskRCNN.
    (a) The decoder is exactly the same as MaskRCNN, except that, before region proposal, we augment the FPN feature maps with a latent code.
    (b) The encoder takes the (non-augmented) FPN feature maps and a list of ground-truth instances and predicts a diagonal Gaussian distribution over latent codes.
    (c) The prior takes the (non-augmented) FPN feature maps and predicts a diagonal Gaussian distribution over latent codes.
    }
    \label{fig:detailed_arch}
\end{figure}

In this section, we explain the architecture of Latent-MaskRCNN (Section~\ref{sec:latentmrcnn}) in more detail.
See Figure~\ref{fig:detailed_arch} for a visual overview.

In our experiments, we use latent codes of a fixed vector dimensionality $z \in \mathbb{R}^d$.
We found that $d = 64$ was a reasonable choice that worked for all datasets.

In the decoder (Figure~\ref{fig:detailed_arch} (a)), we tile the latent codes across the spatial dimensions of the image and concatenate them with the feature maps coming out of the FPN. Then we use a few convolutional layers to project the combined feature maps back down to their original channel dimensionality.
The rest of the model (region-proposal, classifier head, mask head) uses these latent-augmented feature maps but is otherwise unchanged from MaskRCNN.
As we discussed in Section~\ref{sec:latentmrcnn}, this scheme allows the latent codes to influence every stage of the prediction.

The encoder (Figure~\ref{fig:detailed_arch} (b)) produces a Gaussian distribution over latent codes based on both the image and the ground-truth instances.
For each instance $y_i$, we extract RoI-aligned features from the FPN feature maps, as well as of the ground-truth instance masks.
Then we use a small CNN to embed each one into a single feature vector.
At this point, we want to accumulate information from among the per-instance features, and produce a latent representation of a fixed size.
To do this, we employ a graph neural network (GNN, \cite{gnn}), where each instance constitutes a node in a fully-connected graph.
After several graph network layers, we mean-pool across the node features and use a fully-connected layer to produce a mean and log-variance for our latent code.

The prior (Figure~\ref{fig:detailed_arch} (c)) also predicts a Gaussian distribution over latent code, but it does not have access to the ground-truth instances.
Instead, we apply a few convolutional layers to the FPN feature maps, mean-pool across the spatial dimensions, and then predict a mean and log-variance using a small MLP.

Consider the following training objective, and observe that it is equivalent to the standard VAE objective when $\beta = 1$.
$$
\mathcal{L}(x, y) = \mathbb{E}_{z \sim q(z|y,x)}[ \log p(y|x,z) ] - \beta D_{KL}( q(z|y,x) || p(z|x) )
$$
When training VAEs, it is common to use a KL warmup, where $\beta$ is ramped up from 0 over the first few epochs of training.
This helps encourage the latent codes to become more informative since there is no penalty for using them at the beginning of training (when the encoder is untrained and they are uninformative).

We trained all models on a single 8-GPU machine (1080Ti) of our internal cluster. 
Training takes roughly 20-30 hours for each model, depending on the dataset (on par with MaskRCNN). Our code is available at \url{https://segm.yuxuanliu.com}

\begin{figure*}[ht]
    \centering
    \includegraphics[width=0.8\linewidth]{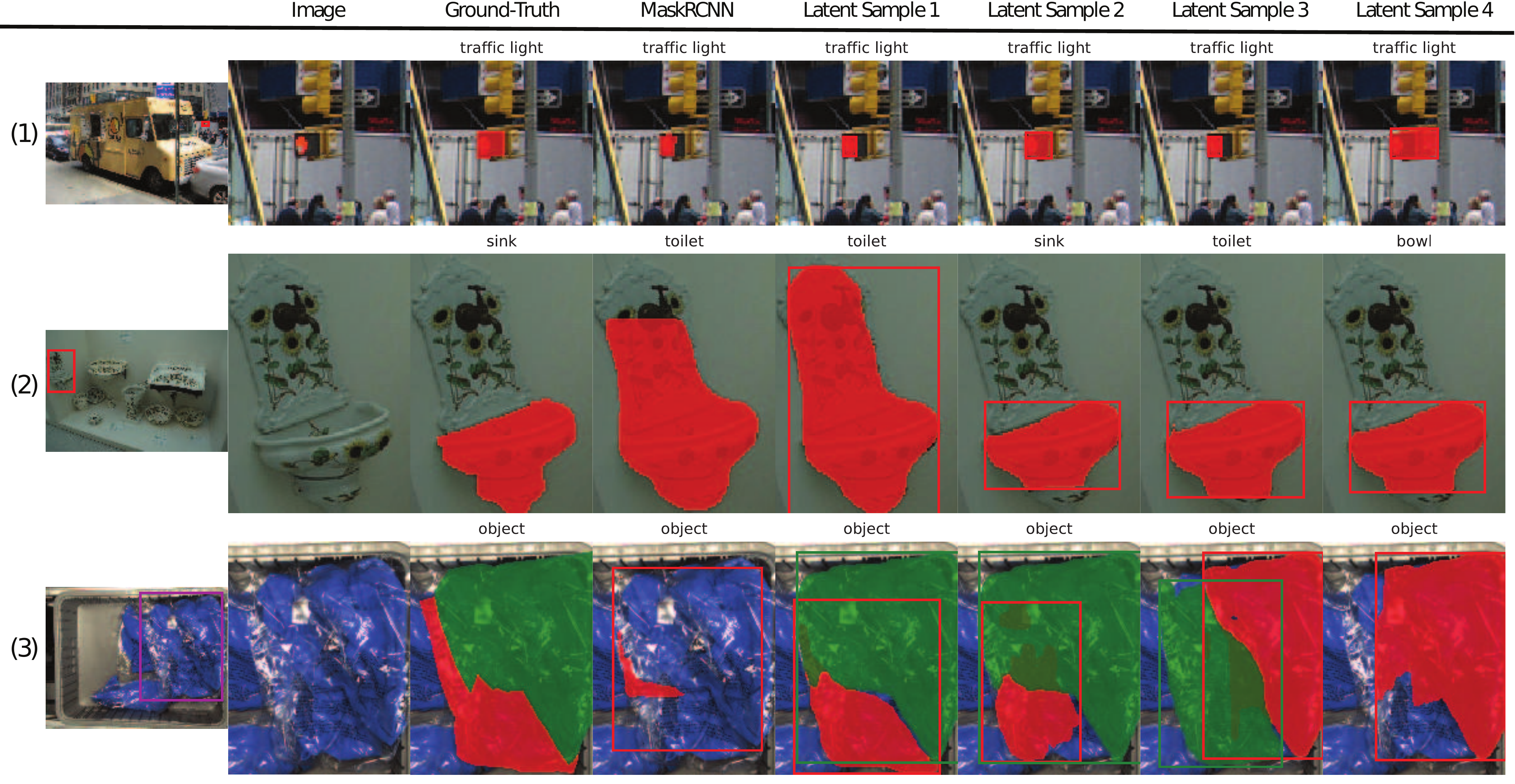}
    \caption{
    Each row corresponds to a single image.
    Columns 1-2 show the original image and a zoomed-in version.
    The remaining columns show instance masks, and the word above the image is the class of the instance.
    Column 3 is the ground truth, column 4 is MaskRCNN's prediction,
    and columns 5-8 are samples from Latent-MaskRCNN.
    }
    \label{fig:samples}
\end{figure*}

\subsection{Qualitative Evaluation of Samples}\label{sec:qualsamples}

In this section, we qualitatively explore what kinds of uncertainty Latent-MaskRCNN can express.
In Figure \ref{fig:samples}, we visualize samples from the model on images from various datasets, and observe that it does capture several distinct types of ambiguity:

\textbf{Category confusion:}
Latent-MaskRCNN can express meaningful uncertainty in the classification head.
In row 2, MaskRCNN confidently classifies this sink as a toilet, while different samples from Latent-MaskRCNN classify it as toilet, sink, and bowl.

\textbf{Category imprecision:}
Even when the class of an instance is obvious, there still may be ambiguity in how the category is defined.
For example, in row 1, both MaskRCNN and Latent-MaskRCNN are (correctly) confident that this instance is a traffic light.
However, depending on how you define exactly what constitutes the extent of the traffic light, the instance mask may look very different.
Latent-MaskRCNN samples a wide range of plausible possibilities, but MaskRCNN picks a single (in this case incorrect) mode.

\textbf{Object and mask ambiguity:}
Latent-MaskRCNN lets us sample from a distribution over sets of objects.
For example, in row 3, we see that the samples contain different numbers of objects, variety in bounding boxes, and variety in instance masks.
Even though none of the samples are perfect, they are all plausible and all markedly better than MaskRCNN.

In row 2, we see examples where Latent-MaskRCNN's hypotheses express uncertainty in both mask and bounding box, while MaskRCNN picks a single mode.

\subsection{Metrics}
\label{appendix:metrics}

In this section, we describe in detail the application-specific metrics that we used in Section~\ref{sec:app_use_cases}.

First, we provide a brief overview of the mAP (mean average precision) metric that is commonly used to evaluate instance segmentation predictions.
We say that a predicted instance and a ground-truth constitute a match if they have the same class, and their masks are within some IoU threshold of each other.
We iterate through the predicted instances in order of decreasing confidence and determine which ones have a matching ground-truth instance (note that each ground-truth instance can only appear in one match).
The predicted instances that appear in a match are considered true positives, and the remaining ones are considered false positives.
We can then plot a precision-recall curve using this information, and compute the area under the curve.
Typically this is done for each class, and we obtain the \textit{average precision} (AP) by averaging across classes.
Note that the AP still depends on the IoU threshold that we choose.
The \textit{mean average precision} (mAP) that is typically reported is the AP averaged across several IoU thresholds ($0.5, 0.55, \dots, 0.9, 0.95$).

mAP is a well-balanced metric, in the sense that it equally penalizes over-segmentation and under-segmentation, and considers both precision and recall.
However, for specific applications like the ones we discussed in Sections~\ref{sec:precision} and \ref{sec:recall}, we may want different metrics that better reflect the asymmetric costs of different types of errors.

For the high-precision use cases like the one discussed in Section~\ref{sec:precision}, we considered the \textit{max recall at high precision} (MR@HP).
We perform the matching procedure in a similar manner to mAP, except that we use IoP instead of IoU.
Next, for a given precision threshold $p_i$ and IoP threshold $\tau_j$, we can compute the recall that each model achieves (or zero, if it never achieves precision $p_i$).
The MR@HP metric is the average of these recalls, over a range of precision threshold $p$ and IoP thresholds $\tau$:
$$
\text{MR@HP} = \frac{1}{|p| \cdot |\tau|}\sum_{p_i \in p, \tau_j \in \tau}\max_{t: \text{Precision}(\tau_j) \ge p_i} \text{Recall}(t, \tau_j)
$$
For the evaluation in Section~\ref{sec:precision}, we use $p = \tau = [0.75, 0.8, 0.85, 0.9, 0.95]$.

AR was introduced in \cite{coco} and is gaining popularity in the instance segmentation community as a complement to mAP.
It also uses the matching procedure that mAP does, but then it simply averages each method's recall over the standard range of IoU thresholds.

For high-recall use cases like the one discussed in Section~\ref{sec:recall}, we considered the average recall (AR), but using IoG instead of IoU.
This measures both recall (did the model predict all the instances?) while also penalizing over-segmentation (are there any predicted masks that are too small).

\subsection{Algorithms}
\label{appendix:algos}

\begin{algorithm}[t]
\label{algo:confidence_mask}
\caption{Confidence Mask}
\textbf{Given:} confidence requirement $p$, $k$ samples of $y$ each with masks $m_j \in y$ \\
 Initialize $M \leftarrow \{\}$ to be a set of predicted masks \\
 \While{masks in $M$ have score $>0.1$}{
    \For{$m_h\in y_i$}{
        Compute the area of the intersection $I_{jg} = |m_h \cap m_j \setminus M|$ with all other masks $m_j \in y_g$\\
        Let $C$ be the set of masks with the $kp$ highest $I_{jg}$ where each mask must come from a unique $y_g$\\
        Compute the intersection $m^*_h = \bigcap_{m\in C}m \setminus M$ ignoring predicted masks $M$ 
    }
    Add mask with highest score $\arg \max s_{m^*_h}$ to $M$
 }
\end{algorithm}

\begin{algorithm}[t]
\label{algo:union_nms}
\caption{Union-NMS}
\textbf{Given:} Masks $M = \{m_i\}_{i=1}^{n}$ sorted by confidence, IoU threshold $\tau$ \\
Initialize $S \leftarrow \{\}$ to an empty map \\
\For{$i = 1:n$}{
    \If{$i \in K$}{
        \textbf{continue}
    }
    $S[i] \leftarrow \{\}$ \\
    \For{$j = i+1:n$}{
        \If{ IoU$(m_i, m_j) > \tau$ }{
            Add $j$ to $S[i]$
        }
    }
}
Initialize $U \leftarrow \{\}$ \\
\For{$i \in \text{keys}(S)$}{
    $m \leftarrow m_i$ \\
    \For{$j \in S[i]$}{
        $m \leftarrow m \cup m_j$
    }
    Add $m$ to $U$
}
\textbf{Result:} $U$

\end{algorithm}

In Section~\ref{sec:app_use_cases}, we proposed Confidence Masks and Union NMS as two applications of Latent-MaskRCNN. Algorithm~\ref{algo:confidence_mask} details the iterative greedy confidence mask prediction, and Algorithm~\ref{algo:union_nms} details the Union NMS procedure.

 \subsection{Uncertainty}

\begin{table}[t]
  \centering
  \caption{ROC AUC using scores to predict ground truth IoU $> 0.5$}\label{table:roc}
  \begin{tabular}{|c|c|c|c|}
    \hline
  Method&COCO &Cityscapes  &Apparel-5k\\
    \hline
    MaskRCNN & 0.7372 & 0.8443 & 0.7756\\
    Confidence Mask & \textbf{0.7941} & \textbf{0.8666} & \textbf{0.9435}\\
    \hline
  \end{tabular}
\end{table}

  \begin{figure}[t]
   \centering
   \includegraphics[width=0.5\linewidth]{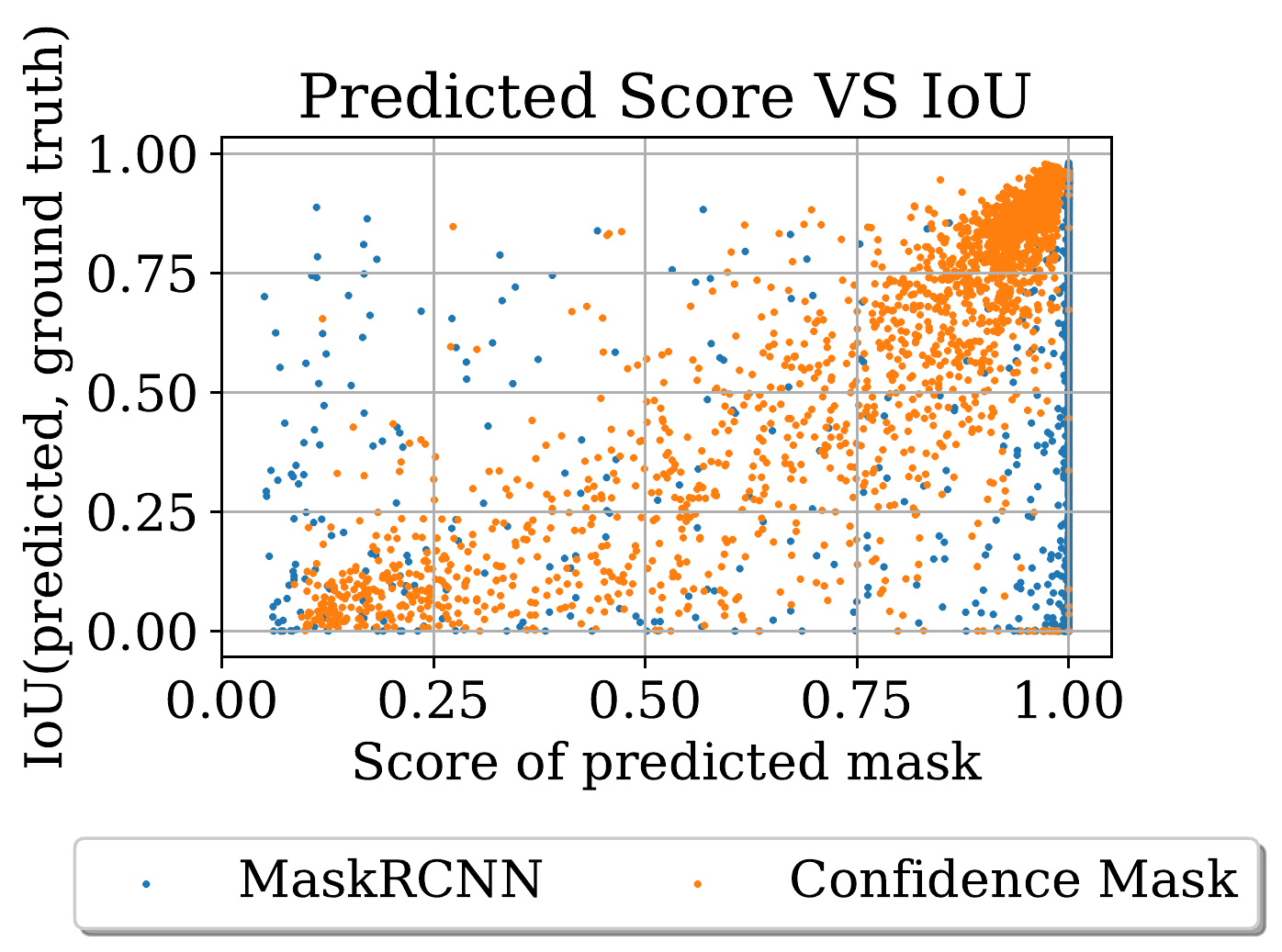}\includegraphics[width=0.5\linewidth]{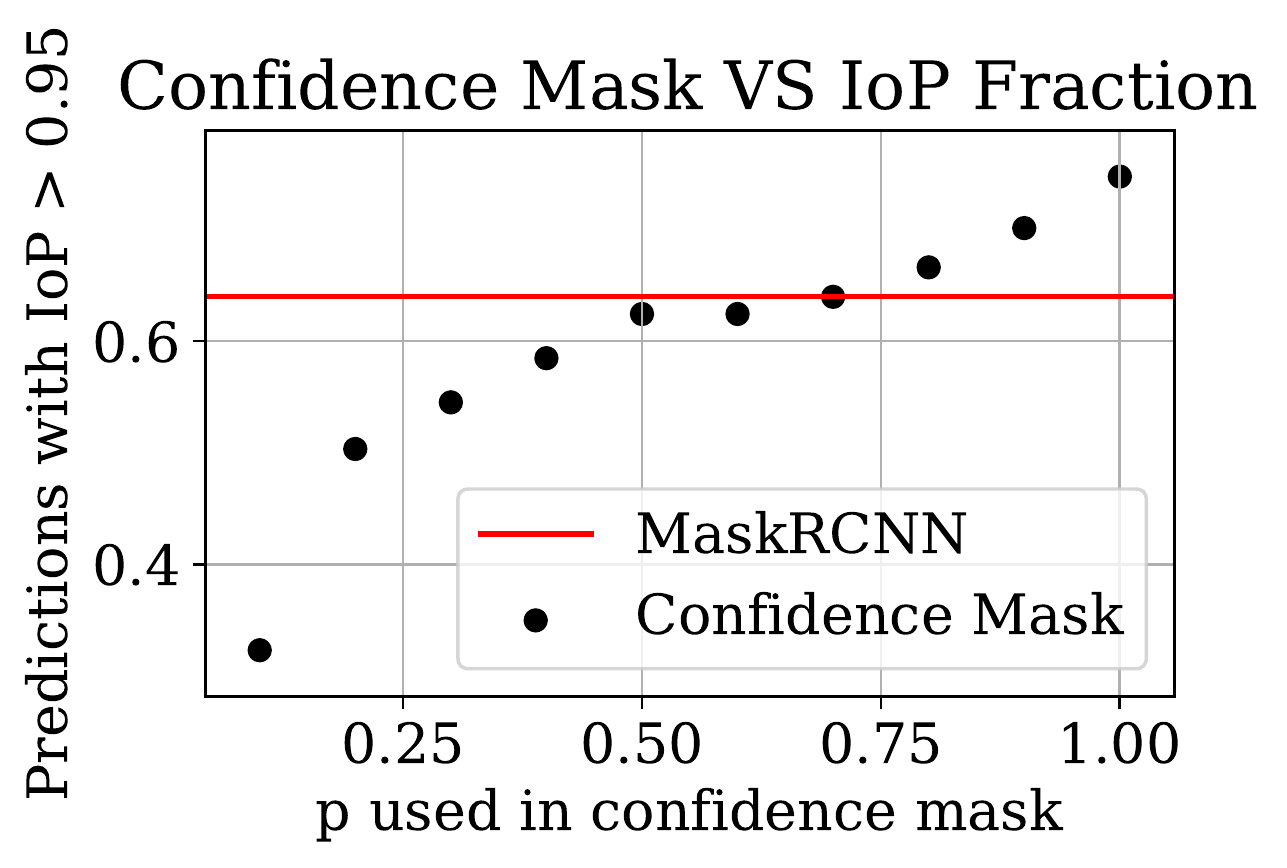}
   \caption{Left: Correlation of predicted score with ground truth IoU of each mask on the Apparel-5k dataset. Latent confidence mask scores tend to be more correlated while MaskRCNN is overconfident. Right: Increasing the confidence requirement $p$ increases the IoP of the predictions while MaskRCNN can only realize one IoP distribution.}
   \label{fig:confidence_scores}
 \end{figure}

 In Section~\ref{sec:score_conf}, we proposed a scoring method for confidence mask predictions.
 Existing models such as MaskRCNN tend to be extremely confident in general, even when they are wrong.
 However, a distributionally-expressive model should be able to express more calibrated confidence scores.

 To evaluate whether these scores correlate well with mask quality, we compare the computed score with the IoU between the predicted mask and the closest ground-truth mask (which indicates how correct the mask is).
 Figure~\ref{fig:confidence_scores} plots these two quantities on the Apparel-5k dataset.
 We see that MaskRCNN almost always has full confidence (regardless of the IoU), while the Latent-MaskRCNN produces confidence scores that are reasonably correlated with the mask accuracy.
 
 Table~\ref{table:roc} also shows ROC AUC where we use each model's score to predict when the ground truth IoU exceeds some threshold $\text{IoU}(c_p, G) \ge 0.5$.
 Across all three datasets, Latent-MaskRCNN has higher AUCs compared to MaskRCNN, suggesting its scores are more predictive of mask quality. 
 This suggests that a distributionally expressive model like Latent-MaskRCNN with the proposed confidence mask scoring can be more effective at measuring uncertainty.

When we proposed $p$-confidence masks in Section~\ref{sec:precision}, we wanted each confidence mask to be contained within some object with probability $p$. To evaluate this, we can compare the fraction of predictions made by each confidence mask that have IoP $> 0.95$ with the confidence requirement $p$. Figure~\ref{fig:confidence_scores} shows the IoP quantiles for different confidence mask $p$'s on Apparel-5k. We generally find that increasing the confidence requirement, $p$, also increases the IoP of the predictions. Meanwhile, MaskRCNN can only achieve one distribution of IoP's since no knobs control the confidence of its outputs.

\end{document}